\documentclass[twocolumn,conference]{IEEEtran}
\usepackage[T1]{fontenc}
\usepackage[latin9]{inputenc}
\usepackage{amsmath}
\usepackage{amssymb}
\usepackage{graphicx}
\usepackage[unicode=true,
 bookmarks=true,bookmarksnumbered=true,bookmarksopen=true,bookmarksopenlevel=1,
 breaklinks=false,pdfborder={0 0 0},pdfborderstyle={},backref=false,colorlinks=false]
 {hyperref}
\hypersetup{pdftitle={Your Title},
 pdfauthor={Your Name},
 pdfpagelayout=OneColumn, pdfnewwindow=true, pdfstartview=XYZ, plainpages=false}

\makeatletter
\usepackage[caption=false,font=footnotesize]{subfig}

\@ifundefined{showcaptionsetup}{}{%
 \PassOptionsToPackage{caption=false}{subfig}}
\usepackage{subfig}
\makeatother

\begin{document}
\title{\thispagestyle{empty}\pagestyle{empty}Swimming locomotion of Soft
Robotic Snakes}
\author{Isuru~S.~Godage, \emph{Member, IEEE}.}
\maketitle
\begin{abstract}
Bioinspired snake robotics has been a highly active area of research
over the years and resulted in many prototypes. Much of these prototypes
takes the form of serially jointed-rigid bodies. The emergence of
soft robotics contributed to a new type of snake robots made from
compliant and structurally deformable modules. Leveraging the controllable
large bending, these robots can naturally generate various snake locomotion
gaits. Here, we investigate the swimming locomotion of soft robotic
snakes. A numerically efficient dynamic model of the robot is first
derived. Then, a distributed contact modal is augmented to incorporate
hydrodynamic forces. The model is then numerically tested to identify
the optimal bending propagation for efficient swimming. Results show
that the soft robotic snakes have high potential to be used in marine
applications.
\end{abstract}

\section{Introduction\label{sec:Introduction}}

Snakes are a remarkable evolutionary success story with a vast range
of habitat including inhospitable deserts, thick tropical forests,
and inhospitable marshes. One key feature to their ability to survive
under challenging terrains is its physical structure; a long, high
degrees of freedom (DoF) slender body which enables snakes to successfully
negotiate countless environmental challenges such as climbing trees,
swimming in the water and marshlands, bury under sand for hunting,
etc. While it is remarkable that they are well adapted to their natural
habitat with a body that lacks sophisticated appendages such as hands
of legs. Instead, the snakes use their bodies as multifunctional appendages
for various situations; propeller to locomote or gripper for crush
prey. Further, the high DoF their bodies facilitate not only the aforementioned
tasks but also enable a range of locomotion gaits such as serpentine,
rectilinear, concertina motions. The few DoF of prior rigid-bodied
snake robots had difficulties in replicating these locomotion gaits
in high fidelity. Soft robotic snakes \cite{key-1,key-7,key-8}, due
to their smooth bending, on the other hand, can assume such \textquotedblleft organic\textquotedblright{}
bending shapes. Prior work involving soft robotic snakes have considered
various terrestrial locomotion.{\let\thefootnote\relax\footnote{{Isuru~S.~Godage is with the School of Computing, DePaul Univeristy, Chicago 60604, USA, e-mail: \href{mailto:igodage@depaul.edu}{igodage@depaul.edu}.}}}\\{\let\thefootnote\relax\footnote{{This work supported in part by the National Science Foundation grant IIS-1718755.}}}

\section{Methodology\label{sec:Methodology}}

\subsection{Prototype Description\label{subsec:Prototype-Description}}

The prototype soft robot snake is shown in Fig. \ref{fig:prototype}.
It consist of three, serially attached, soft modules each powered
by two pneumatically powered soft McKibben type actuators. These actuators
extends proportionally to supplied input pressure (up to $3\,\text{bars}$)
which are precisely controlled by digital pressure regulators. Similar
to continuum sections reported in \cite{key-2}, these actuators are
bundled together so that, when there is a pressure differential of
a soft module, it causes the soft module to bend. By controlling these
bends synchronously, the snake robot can generate various modes locomotion
gaits.

Each soft actuator has an unactuated length, $L_{i0}=0.15m$ and can
extend by $0.065m$ at $3\,\text{bars}$. Rigid plastic frames (made
of ABS thermoplastic) of $r_{i}=0.0125\,m$ and $2.54\,mm$ thickness
are used to mount the soft actuators and connect adjacent soft modules.
Rigid 3D-printed plastic constrainer plates are used along the length
of soft modules to maintain soft actuators in-parallel to the central
axis of soft modules with a $r_{i}$ clearance from the central axis.
In addition, these constrainer plates provide improved strength for
this long and slender soft robot snake to maintain its structural
integrity during locomotion and generate reaction forces required
to locomote. Each soft module, including the soft actuators and plastic
constrainer plates, weigh close to $0.1\,$$kg$.

\subsection{System Model\label{subsec:System-Model}}

\begin{figure}[t]
\subfloat[\label{fig:prototype}]{\includegraphics[height=2.4cm]{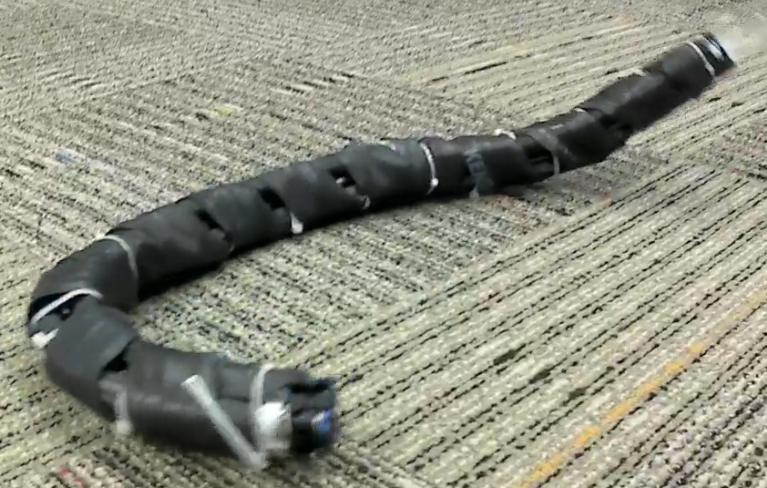}

}\subfloat[\label{fig:Schematical-illustration-of}]{\begin{raggedright}
\includegraphics[height=2.4cm]{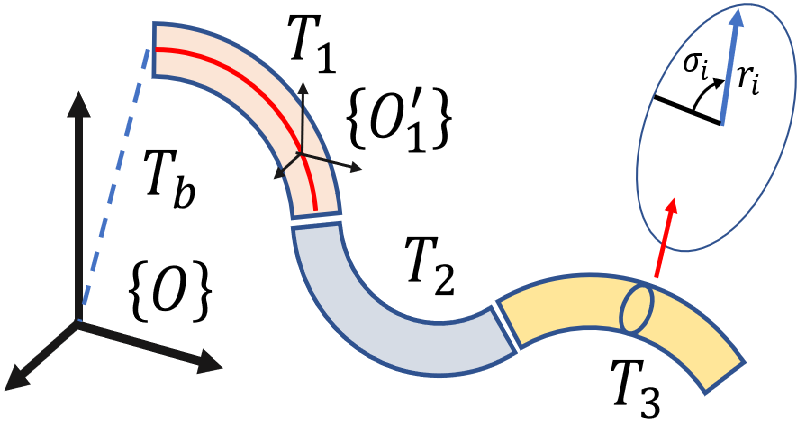}
\par\end{raggedright}
}

\caption{(a) Soft robot snake prototype crawling on a carpeted floor, (b) Schematical
illustration of the snake robot. Each soft section is enumerated while
the base of the floating coordinate system is attached tot he base
of the first soft section. The homogeneous transformation matrices
(HTM) for the floating base and each soft section are also shown.
The moving coordinate system $\left\{ O_{1}^{\prime}\right\} $ is
shown for the first soft section.}

\end{figure}

\begin{center}
Consider the schematic of any $i^{th}$ soft robot module of the snake
as shown in Fig. \ref{fig:Schematical-illustration-of}. It consists
of three mechanically identical variable length actuators with $L_{i0}\in\mathbb{R}$
and $l_{ij}\left(t\right)\in\mathbb{R}$, where $j\in\left\{ 1,2,3\right\} $
and $t$ is the time. Hence the length of an actuator at any time
is $L_{ij}=L_{i0}+l_{ij}(t)$. 
\par\end{center}

\subsubsection{Kinematic Model\label{subsec:Kinematic-Model}}

The kinematics of a snake robot can be formulated by extending the
modal kinematics proposed by the author in \cite{key-3}. Unlike the
manipulator case, however, the model should be capable of generating
points on the skin of the soft robot to incorporate contact dynamics
in Section \ref{subsec:Contact-Dynamics}. Let the joint space vector
of any $i^{th}$ soft robotic section be $\boldsymbol{q}_{i}=\left[l_{i1}\left(t\right),\,l_{i2}\left(t\right)\right]^{T}$
where $i\in\left\{ 1,2,3\right\} $ is the section number and $t$
is the time. Utilizing the results from \cite{key-3}, we can derive
the homogeneous transformation matrix (HTM) at any point along the
neutral axis of a single soft section, $\mathbf{T}_{i}\in\mathbb{SE}^{3}$,

\begin{align}
\mathbf{T}_{i}\left(\boldsymbol{q_{i}},\xi_{i},\beta_{i}\right) & =\left[\begin{array}{cc}
\mathbf{R}_{i}\left(q_{i},\xi_{i}\right) & \boldsymbol{p}_{i}\left(q_{i},\xi_{i}\right)\\
\mathit{\boldsymbol{0}} & 1
\end{array}\right]\cdots\nonumber \\
 & \qquad\left[\begin{array}{cc}
\mathbf{R}_{z}\left(\sigma_{i}\right) & \mathcal{\mathit{\boldsymbol{0}}}\\
\mathit{\boldsymbol{0}} & 1
\end{array}\right]\left[\begin{array}{cc}
\boldsymbol{0} & \boldsymbol{p}_{x}\left(r_{i}\right)\\
\mathit{\boldsymbol{0}} & 1
\end{array}\right]\label{eq:ith_kin}
\end{align}
where $\mathbf{R}_{i}\in\mathbb{SO}^{3}$ is the rotational matrix,
$\boldsymbol{p}_{i}\!\in\mathbb{R}^{3}$ is the position vector, and
$\xi_{i}\in\left[0,1\right]$ is a scalar to define points along the
soft robotic section with 0, and 1 denotes the base and the tip of
the module. In addition to the previous results in {[}{]}, note that
we introduce two HTMs with $\mathbf{R}_{z}\in\mathbb{SO}^{3}$ is
the rotation matrix about the $+Z$ axis and $\boldsymbol{p}_{x}\in\mathbb{R}^{3}$
is the translation matrix along the $+X$ of $\left\{ O_{i}^{\prime}\right\} $
where $\sigma_{i}\in\left[0,2\pi\right]$. The latter HTM defines
the physical outer boundary (skin) of the soft robotic sections such
that the contact dynamics could be implemented.

Utilizing \eqref{eq:ith_kin} with a floating coordinate system, $\mathbf{T}_{b}\in\mathbb{SE}^{3}$,
the complete kinematic model along the body of the snake robot is
given by
\begin{align}
\mathbf{T}\left(\boldsymbol{q}_{b},\boldsymbol{q},\boldsymbol{\xi}\right) & =\mathbf{T}_{b}\left(\boldsymbol{q}_{b}\right)\prod_{i=1}^{3}\mathbf{T}_{i}\label{eq:complete_kin}
\end{align}
where $\boldsymbol{q}_{b}=\left[x_{b},y_{b},z_{b},\alpha,\beta,\gamma\right]$
are the parameters of the floating coordinate system with $\left[x_{b},y_{b},z_{b}\right]$
denote the translation and $\left[\alpha,\beta,\gamma\right]$ denote
the Euler angle offset of the base of the module 1 with respect to
$\left\{ O\right\} $. The composite vector $\boldsymbol{q}=[\boldsymbol{q}_{1},\boldsymbol{q}_{2,},\boldsymbol{q}_{3}]\in\mathbb{R}^{9}$
and $\boldsymbol{\xi}=\left[0,3\right]\in\mathbb{R}$.

\subsubsection{Contact Dynamics\label{subsec:Contact-Dynamics}}

The skin or the outer layer of the soft modules of the snake robot
makes contact with the environment to generate locomotion. Given the
smooth deformation and continuous nature of the skin, in reality,
the contact surface is a continuum. However, implementing such contact
dynamics not only mathematically challenging but also computationally
inefficient. To overcome this challenge, we discretize the parameters
$\boldsymbol{\xi}$ (31 points in $\left[0,3\right]$) and $\sigma_{i}$
(10 points in $\left[0,2\pi\right]$) of \eqref{eq:complete_kin},
define a discretized set of 310 total contact points on the skin of
the robot snake. The reaction forces of those points are then computed
during actuation and included in the complete dynamic model. Without
losing generality, let the reaction force at any contact point (defined
by $\xi_{j}$ and $\sigma_{k}$), $S_{jk},$be

\begin{align}
S_{jk} & =\frac{1}{2}C_{D}\rho A\upsilon_{jk}^{2}\label{eq:reaction_force}
\end{align}
where $C_{D}$is the drag coefficient of water, $\rho$ is the density
of water, $A$ is the area of contact point (we assume $1\,cm^{2}$area
at the contact point), and $\upsilon_{jk}$ is the velocity at the
contact point with respect to $\left\{ O\right\} $. Here, we assume
that the robot lightweight and thus does not submerge in the water.
This is a reasonable assumption because the robot is constructed from
light weight material and is actuated by soft pneumatic muscle actuators.
To enforce this assumption, we add a balancing force by means of a
spring-damper contact model \cite{key-4} to counter the effect of
gravity when the robot touches the water.

\subsubsection{Dynamic Model\label{subsec:Dynamic-Model}}

We extend the recursive integral Lagrangian approach and include floating
coordinate parameters to formulating the dynamics of the soft robot
snake. The equations of motion employed in the simulations is given
by

\begin{align}
\mathbf{M}\ddot{\boldsymbol{Q}}+\mathbf{C}\dot{\boldsymbol{Q}}+\mathbf{D}\dot{\boldsymbol{Q}}+\boldsymbol{G} & =\left[\begin{array}{c}
0\\
\boldsymbol{\tau}_{e}
\end{array}\right]+\sum_{j\in\xi,k\in\sigma}\mathbf{J}_{jk}^{T}S_{jk}\label{eq:simulatino_EoM}
\end{align}
where $Q=\left[\boldsymbol{q}_{b},\boldsymbol{q}\right]\in\mathbb{R}^{12}$
complete floating base jointspace vector, $\mathbf{M}$ is the generalized
inertia matrix, $\mathbf{C}$ is the centrifugal and Coriolis force
matrix, $\boldsymbol{G}$ is the conservative force matrix, $\mathbf{D}=\eta\mathbf{I}\in\mathbb{R}^{6\times6}$
is the damping forces matrix, $\mathbf{J}_{jk}$ is the Jacobian of
$S_{jk}$, and $\tau_{e}$ is the input force matrix (pressure supplied
to soft actuators).

These elastic stiffness limiting values and damping coefficients were
approximately identified by following an experimental procedure similar
to the method proposed in \cite{key-6}. This stiffness value (rounded
to the nearest 100) of soft actuators used in the simulations is $1900Nm^{-1}$and
the damping coefficient (rounded to the nearest 10) is $90Nm^{-1}s$.
The gravitational acceleration used in the simulations is $9.81ms^{-2}$.
Note that the hysteretic effects are negligible relative to the damping
effects and are not considered in this experiment.

The numerical model was implemented and simulations were carried out
in MATLAB 2018a programming environment. We use the variable-step
ODE15s routine for solving \eqref{eq:simulatino_EoM} due to its speed
in handling complex and high-DoF dynamic systems such as the one discussed
in this paper. The simulation data was recorded at 30 samples per
second to preserve the smoothness of jointspace variable changes.
In addition, it facilitates the simulation movie creation at 30 frames
per second.

\begin{figure}[tbh]
\begin{centering}
\includegraphics[width=1\columnwidth]{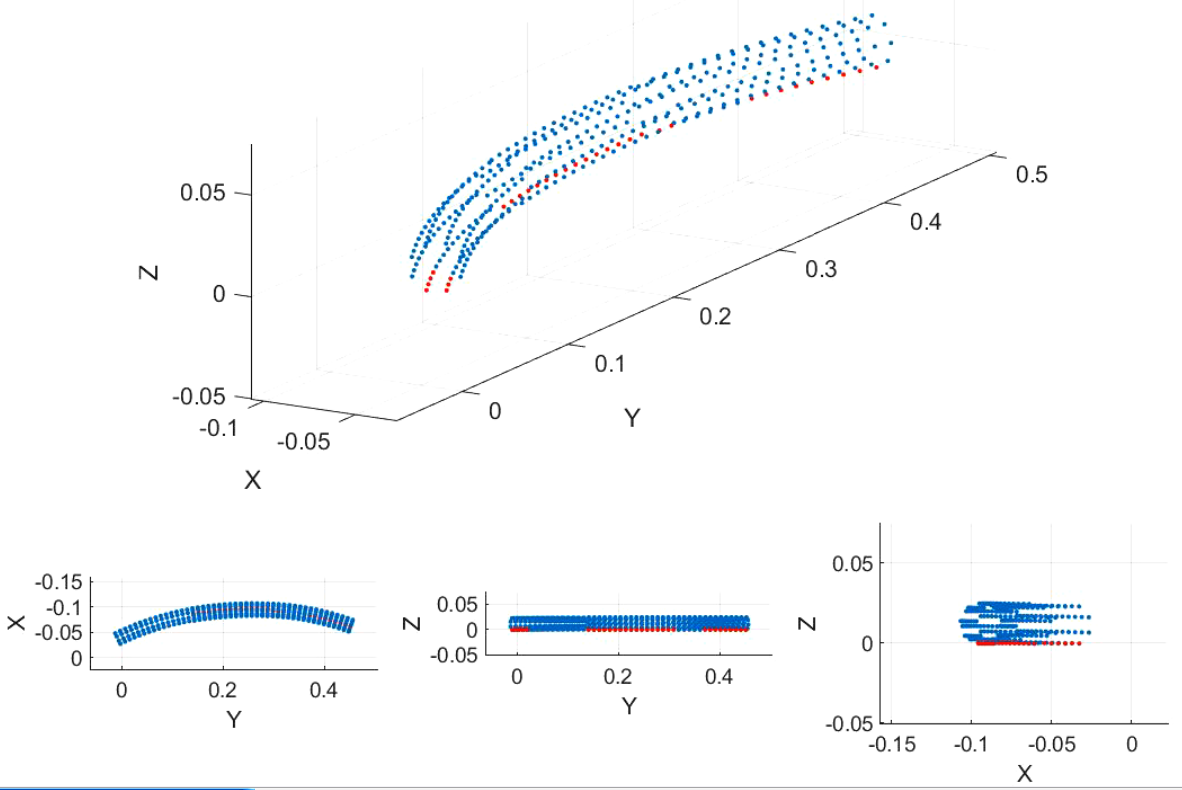}
\par\end{centering}
\caption{Grided contact points on the soft robot snake highlighting the contact
points upon contact with water (red-colored points) where the water
level is at $z=0$. The figures below show the robot when looking
from different perspectives.\label{fig:contact_dyna}}
\end{figure}

\section{Results\label{sec:Results}}

Three numerical studies were conducted to assess the modeling framework
of the soft robotic snake. In the first study, we evaluated the effectiveness
of the discretized contact dynamics. Therein, we simulated the drop
of our robot from $30cm$ height in its unactuated pose (all jointspace
displacements, velocities, and input forces are at 0) to a water surface
simulated at $z=0$ (see Fig. \ref{fig:contact_dyna}). This causes
the robot to accelerate toward the water surface under the influence
of gravity. Upon contacting the water surface, i.e., the discretized
contact points register a negative $z$ coordinate, the spring-damper
contact model activates, and reaction forces act on those points in
the opposite direction to gravity. The model comes to an equilibrium
when the reaction force equals the weight of the robot. The contact
points in red in Fig. \ref{fig:contact_dyna} denotes the points with
active reaction forces while the subfigures below the main figure
show the orthogonal perspective of the same for better visualization.
It should be noted that, while this simplified contact model simulates
the floating of the soft snake robot in water, the contact model makes
uneven contacts (see bottom-center plot), which is not realistic.
Our ongoing work investigates the use of volume-based contact model,
which can accommodate the fluid displacement toward a more realistic
simulation. 

\begin{figure}[tbh]
\begin{centering}
\subfloat[\label{fig:gait_1}]{\begin{centering}
\includegraphics[width=1\columnwidth]{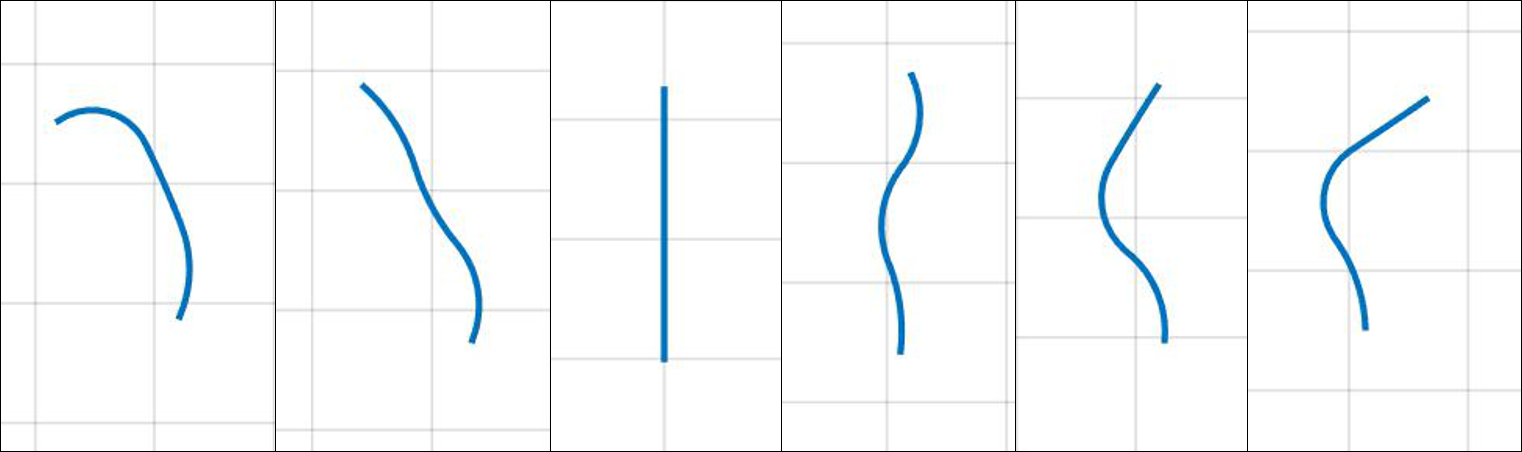}
\par\end{centering}
}
\par\end{centering}
\begin{centering}
\subfloat[\label{fig:gait_2}]{\begin{centering}
\includegraphics[width=1\columnwidth]{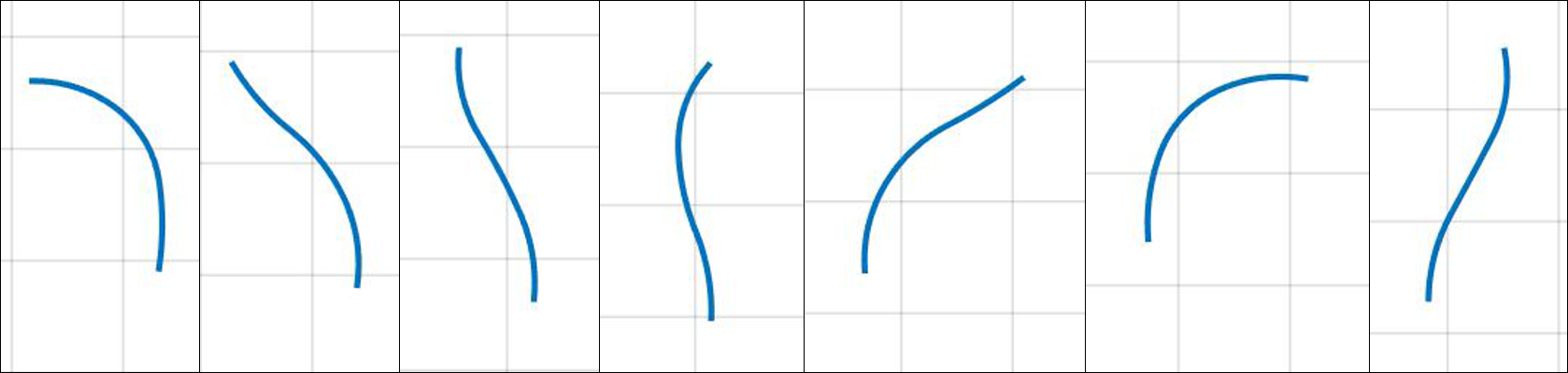}
\par\end{centering}
}
\par\end{centering}
\caption{The snake swimming locomotion resulting from the input signals given
by (a) \eqref{eq:gait_1}, (b) \eqref{eq:gait_2}.}

\end{figure}

The next two studies consider two modes of serpentine locomotion inspired
swimming locomotion of a soft snake robot. Here, we apply the pressure
input signal given by \eqref{eq:gait_1} where $j$ stands for the
soft module index number and $k$ stands for the actuator number.
The period of the gait is $\frac{1}{2}rads^{-1}$ with phase offsets
of multiple of $\frac{\pi}{8}$ added to subsequent soft modules.
The $\pi$ phase shift between the soft actuators of the same soft
modules ensures antagonistic actuation for periodic bidirectional
bending. The pressure signal causes the robot to move with respect
to water, which then causes reactive drag forces at the contact points
noted in \eqref{eq:reaction_force}. The net force of these reaction
forces then causes the robot to move forward. Figure \ref{fig:gait_1}
shows various stages during this locomotion. In the next study, the
phase offsets of multiples of $\frac{\pi}{3}$ are added to the soft
sections and is given by \eqref{eq:gait_2}. This causes the serpentine
gait with larger wave amplitude. Despite resulting in more become
more pronounced wave, the locomotion speed is less than that of the
gait resulted from \eqref{eq:gait_1}. The ongoing work investigates
a systematic methodology for finding the optimal serpentine wave signals
for soft robot snakes as well as implementing other common locomotion
gaits such as concertina and side-winding.

\begin{align}
P_{jk} & =1.5\left[1+\sin\left\{ \frac{1}{2}t+\left(k-1\right)\pi+\left(j-1\right)\frac{\pi}{8}\right\} \right]\label{eq:gait_1}\\
P_{jk} & =1.5\left[1+\sin\left\{ \frac{1}{2}t+\left(k-1\right)\pi+\left(j-1\right)\frac{\pi}{3}\right\} \right]\label{eq:gait_2}
\end{align}

\section{Conclusions}

Snakes are a remarkable evolutionary success story with a vast range
of habitat including inhospitable deserts, thick tropical forests,
and inhospitable marshes. Many snake robotic prototypes to date use
rigid-robotic structures. The recent emergence of soft robotics opens
up a new generation of snake robots based on smooth and continuously
deformable actuators to achieve life-like snake locomotion gaits.
In this work, we introduced a soft robotic snake and its kinematic
and dynamic model formulation with a simplified, point-based contact
dynamics framework to accommodate drag forces for swimming. The model
was implemented on MATLAB and validated its contact dynamics. Input
pressure signals were applied to the model to generate serpentine
swimming locomotion, and results were included. The ongoing work investigates
the use of volume-based contact model for modeling realistic buoyancy
and reactive drag forces.

\end{document}